\pdfoutput=1
\documentclass{bmvc2k}

\usepackage{graphicx}
\usepackage{amsmath}
\usepackage{amssymb}
\usepackage{booktabs}

\usepackage{comment}
\usepackage{overpic}
\usepackage{textpos}
\usepackage{subfig}
\usepackage[capitalize,noabbrev]{cleveref}

\usepackage[font={color=bmv@captioncolor}]{caption}

\graphicspath{{img/}}

\usepackage{floatrow}
\setkeys{Gin}{width=\linewidth}

\usepackage{array}
\newcommand{\tablestyle}[2]{\setlength{\tabcolsep}{#1}\renewcommand{\arraystretch}{#2}\centering\footnotesize}
\newlength\savewidth\newcommand\shline{\noalign{\global\savewidth\arrayrulewidth
  \global\arrayrulewidth 1pt}\hline\noalign{\global\arrayrulewidth\savewidth}}
\newcolumntype{x}[1]{>{\centering\arraybackslash}p{#1pt}}
\newcolumntype{y}[1]{>{\raggedright\arraybackslash}p{#1pt}}
\newcolumntype{z}[1]{>{\raggedleft\arraybackslash}p{#1pt}}
\definecolor{deemph}{gray}{0.6}
\newcommand{\gc}[1]{\textcolor{deemph}{#1}}

\makeatletter
\DeclareRobustCommand\onedot{\futurelet\@let@token\@onedot}
\def\@onedot{\ifx\@let@token.\else.\null\fi\xspace}

\def\eg{\emph{e.g}\onedot}

\def\etal{\emph{et al}\onedot}

\definecolor{nice-green}{RGB}{0, 100, 0}

\title{FitCLIP: Refining Large-Scale Pretrained Image-Text Models for Zero-Shot Video Understanding Tasks}

\addauthor{Santiago Castro*}{https://santi.uy/}{1}
\addauthor{Fabian Caba Heilbron}{https://fabiancaba.com/}{2}

\addinstitution{University of Michigan}
\addinstitution{Adobe Research}

\runninghead{Castro and Caba Heilbron}{FitCLIP}

\def\eg{\emph{e.g}\bmvaOneDot}

\def\etal{\emph{et al}\bmvaOneDot}

\begin{document}

\maketitle

\renewcommand*{\thefootnote}{\fnsymbol{footnote}}
\footnotetext[1]{Work done as an intern at Adobe Research.}
\renewcommand*{\thefootnote}{\arabic{footnote}}
\setcounter{footnote}{0}

\begin{abstract}
Large-scale pretrained image-text models have shown incredible zero-shot performance in a handful of tasks, including video ones such as action recognition and text-to-video retrieval. However, these models have not been adapted to video, mainly because they do not account for the time dimension but also because video frames are different from the typical images (\eg{}, containing motion blur, less sharpness). In this paper, we present a fine-tuning strategy to refine these large-scale pretrained image-text models for zero-shot video understanding tasks. We show that by carefully adapting these models we obtain considerable improvements on two zero-shot Action Recognition tasks and three zero-shot Text-to-video Retrieval tasks. The code is available at \url{https://github.com/bryant1410/fitclip}
\end{abstract}

\section{Introduction}

Imagine is winter season and our quest is to develop an auto-tagging system that recognizes all the activities in our winter vacation footage. Luckily, there have been tremendous advances in the action recognition community \cite{densetraj,i3d,vivit}. For instance, we could leverage one of the existing models that recognize up to 700 human actions \cite{movinets}.  Sadly, it turns out that our family's favorite activity, sledding, is not on the list of categories that these models can recognize. In a traditional supervised setting, we would have to collect many sledding examples to train a new model.  Such a process is labor-intensive, costly to create, and difficult to scale to recognize further new activities. Instead, zero-shot models \cite{larochelle2008zero,NIPS2013_2d6cc4b2,brattoli2020rethinking} can alleviate such a burden by enabling recognition of unseen concepts.

Large pre-trained image-text models, such as CLIP~\cite{clip} and ALIGN~\cite{align}, have shown outstanding zero-shot capabilities on a handful of visual tasks, including video tasks such as Action Recognition and Text-to-Video Retrieval. Such models have overcome the limitations of traditional zero-shot learning algorithms by using abundant images (on the internet) with (free) natural language supervision. Despite their remarkable zero-shot performance in video tasks, there is room for improvement to close the image-to-video domain gap. For instance, recent studies have shown that fine-tuning CLIP yields significant improvements in target video tasks \cite{clip4clip,actionclip}. Unfortunately, fine-tuning and improving performance in a target dataset comes with a cost: harshly penalizing the model's zero-shot capabilities \cite{wortsman2021robust}.

There have been multiple efforts to train video-language models that can be employed for various downstream video understanding tasks. Even though these approaches use video data, their zero-shot capabilities remain poor compared to those exhibited by CLIP~\cite{clip}. It would be unfair not to mention that video-language pretraining methods either train with clean yet two orders of magnitude smaller datasets \cite{frozen_in_time}, or with large datasets with unaligned natural language supervision \cite{howto100m}. The alternative is to scale up further the amount of unaligned natural language supervision abundant on internet videos. In comparison, ALIGN~\cite{align} (in the image space) has shown the ability to cope with noisy supervision by scaling up to the billion-samples scale. However, replicating such experiments with video data would only be possible for selected (if any) industrial players.

This work introduces FitCLIP, a fine-tuning strategy to adapt large-scale image-text pre-trained models for zero-shot video understanding tasks. The goal of FitCLIP is to retain the knowledge of CLIP~\cite{clip} while gently adapting and learning how video data looks. Our method leverages relatively small labeled and extensive pseudo-labeled video data to train a student network. To validate the effectiveness of FitCLIP, we designed and set zero-shot benchmarks for two popular video understanding tasks: action recognition and text-to-video retrieval. Our experiments empirically validate the effectiveness of distillation to better train and fine-tune multimodal video models and show that FitCLIP establishes a new state-of-the-art for zero-shot video recognition and retrieval. Our design incorporates weight-space ensembling in a strategic manner, which has not been explored before, as far as we know.

\noindent\textbf{Contributions.} Our key idea is to develop a method to refine large-scale pretrained image-language models to zero-shot video use-cases. Our work brings two contributions:\\
\textbf{(1)} We introduce FitCLIP, a refinement strategy, and model for zero-shot video understanding. The model leverages abundant knowledge in large-scale image models and a distillation strategy to learn \textit{new} video knowledge. We describe FitCLIP in (\cref{sec:method}).\\
\textbf{(2)} We evaluate FitCLIP and competitive baselines in a newly designed zero-shot benchmark (\cref{sec:benchmark}). Our experiments include results for two sets of video understanding tasks, action recognition, and text-to-video retrieval, where we show the value of FitCLIP (\cref{sec:experiments}).

\section{Related Work}

\noindent\textbf{Zero-shot Video Understanding.} Multiple zero-shot methods have been proposed to tackle popular tasks such as action recognition \cite{brattoli2020rethinking,chen2021elaborative}, text-to-video retrieval \cite{videoclip}, and localization-related tasks \cite{jain2015objects2action,zhang2020zstad}. Most of the zero-shot action recognition literature either follows an attribute-based approach or leverages word embedding to transfer knowledge \cite{liu2011recognizing,jain2015objects2action,gan2016recognizing,gao2019know,mandal2019out,brattoli2020rethinking,chen2021elaborative}.  Differently, in the text-to-video retrieval task, zero-shot methods leverage large-scale natural language supervision to pre-train video-language models. After pretraining, these models can then be employed and tested in text-to-video retrieval tasks. Similar to \cite{frozen_in_time,videoclip}, our work leverages natural language supervision from video titles to unlock zero-shot capabilities. However, we focus on adapting already well-trained image-text models to videos rather than learning a video-language model from scratch.

One of our goals is to establish a benchmark for zero-shot action recognition and text-to-video retrieval. Previous efforts have devoted insightful analyses to creating \textit{true} zero-shot evaluation for action recognition \cite{Gowda2021}. These efforts are valuable for the traditional zero-shot setting where methods use a close vocabulary of (seen) actions, but they do not fit when zero-shot models learn with natural language supervision. Instead, we follow standard (full) tests on popular action recognition datasets and well-established text-to-video retrieval datasets.

\noindent\textbf{Visual-Language Pretraining.} Pretraining visual models with natural language became a popular learning strategy in the image domain \cite{mori1999image,srivastava2012multimodal,joulin2016learning,Desai_2021_CVPR,clip}. The idea of matching images with text dates back to the late 90s when Mori \etal{} trained models to predict nouns and adjectives from image-text pairs \cite{mori1999image}. Others modernized this idea using large-scale datasets to train CNNs \cite{joulin2016learning}. However, only recently, Radford \etal{} took this idea to the next level \cite{clip}. They trained CLIP, a dual image-text encoder, with more than 400M images and text descriptions using a contrastive objective \cite{infonce}. Our work builds upon CLIP and adapts it to video use-cases while preserving its zero-shot capabilities.

Video-language pretraining also gained traction in the video space. Despite the progress, it has been hard for video-language methods to compete in zero-shot settings with image-language pre-trained models. We argue this is due to the limited availability of videos with clean (and aligned) natural language supervision. For instance, Frozen in Time~\cite{frozen_in_time} trains a transformer-based architecture on the WebVid dataset, which contains 2.5M humanly curated video-title pairs. The dataset is at least two orders of magnitude smaller than the dataset to pre-train CLIP~\cite{clip}. The importance of large and diverse data emerges when we compare Frozen in Time with CLIP in zero-shot video tasks. Others \cite{howto100m,mil_nce} have trained with the relatively larger HowTo100M dataset, which contains 100M unaligned video-text pairs. Still, the zero-shot capabilities of these models remain subpar to what CLIP can provide. Our approach, FitCLIP, leverages the WebVid dataset \cite{frozen_in_time} as a rich source to adapt CLIP for zero-shot video understanding tasks.

\noindent\textbf{Refining Large-scale Image Models.} DistInit~\cite{distinit} explored distilling image models for video. More recently, CLIP's strong visual representation inspired multiple researchers to explore its usage for video tasks \cite{clip_straight,clip4clip,camoe,actionclip,clip,fang2021clip2video}. CLIP4Clip, for instance, proposed a straightforward strategy to fine-tune CLIP for the text-to-video retrieval task \cite{clip4clip}. Surprisingly, their simple method sets a new state-of-the-art in various datasets. Similarly, ActionCLIP introduced a novel paradigm to action recognition harnessing CLIP's general visual knowledge \cite{actionclip}. While existing approaches effectively boost performance on target datasets, and tasks, they have not show to preserve the original CLIP zero-shot capabilities (also based on early experiments we ran).

\section{Method: FitCLIP}%
\label{sec:method}

Our goal is to train a model that \textit{expands and complements} large image-language models \cite{clip,align} for zero-shot video (see \cref{fig:fitclip}). To do so, we introduce FitCLIP, a refinement strategy that leverages small labeled and large pseudo-labeled data together with existing knowledge acquired from large image-text pairs. FitCLIP includes two steps. The first step trains a model, in a Teacher-Student fashion, leveraging both: labeled video-text pairs and pseudo-labels generated by a teacher model. The second step fuses the existing knowledge of the teacher, a large-scale pre-trained image-language model, with the student trained on video data. We call the resulting model the same as our refinement strategy, FitCLIP.

\subsection{Teacher-Student Fine-tuning}

\begin{figure*}
	\centering
	\includegraphics[width=0.95\linewidth]{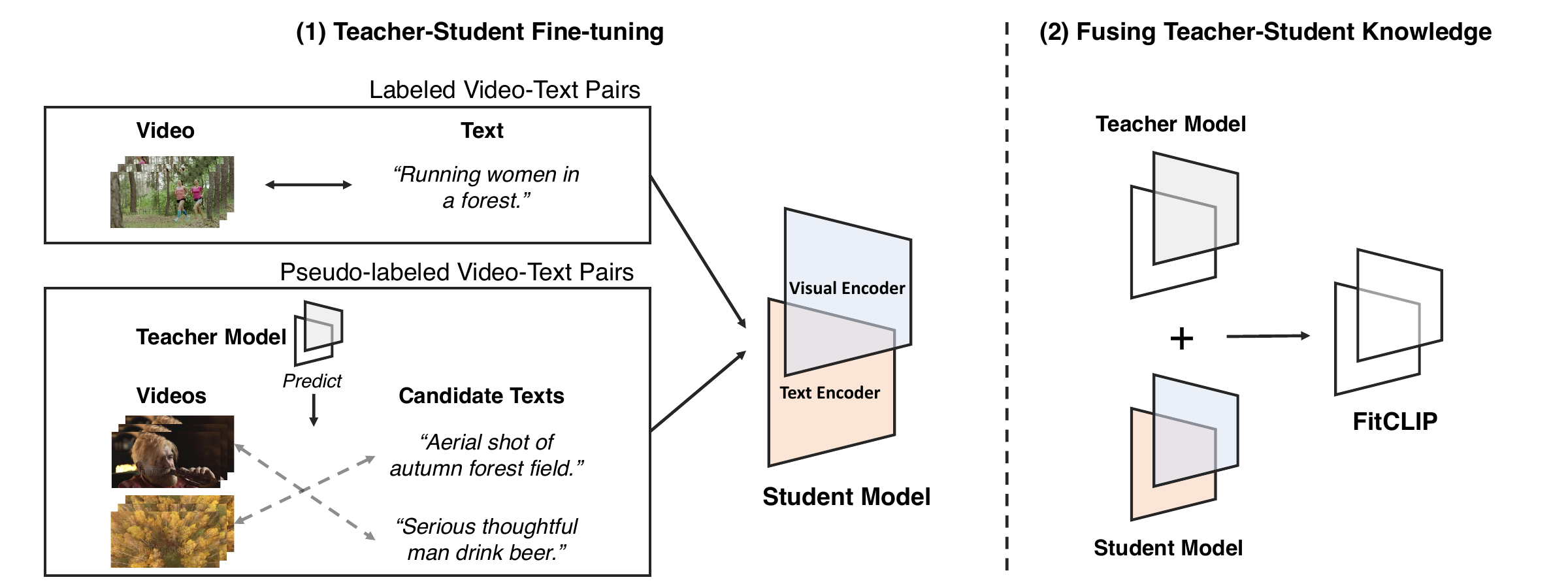}
	\caption{
    	\textbf{FitCLIP refinement strategy and model.} We propose a refinement strategy to adapt large-scale image-text pretrained models. Our strategy first trains a model in a Teacher-Student fashion. To do so, we leverage labeled and pseudo-labeled (with a teacher) video-text pairs. This process, Step (1), yields a Student model that captures video-relevant knowledge while being compatible with the teacher. In step (2), similar to \cite{wortsman2021robust}, we combine the Teacher and Student weights to create our final model, FitCLIP. 
    }
	\label{fig:fitclip}
\end{figure*}

Our goal is to train a model using video-text pairs while leveraging knowledge from image-language representations. One alternative is to reuse image-language encoder weights and fine-tune them in a target dataset \cite{clip4clip,actionclip}. Such an approach is effective in boosting performance for in-distribution datasets but tends to fail at preserving the zero-shot capabilities of the original model's weights due to catastrophic forgetting \cite{french1999catastrophic}. Instead, we focus on gently refining the original image-language model's weights by incorporating a two-fold strategy. We use a small sample of labeled data to avoid model drift \cite{royer2015classifier} (because of using a much smaller batch size and less diverse dataset), and we regularize the learning process by adding pseudo-labels generated with the original image-language model. Note that our strategy shares intuitions with the Knowledge Distillation literature \cite{distillation}, where a Teacher-Student analogy is used to describe the process of training a Student with priors derived from a strong Teacher model. \Cref{fig:fitclip} (step 1) illustrates the process to train our Student model.

\noindent\textbf{Data Subsets.} Our fine-tuning strategy relies upon two subsets of data: a small labeled dataset of video-text pairs and an unlabeled (unaligned) set of video-text candidate pairs. The labeled subset contains a collection of videos matched with one text describing their visual content. These video-text pairs are of high quality and made by a human. The unlabeled subset also contains a list of videos and a list of text descriptions. However, the match between a video and the best describing text does not exist in this subset.

\noindent\textbf{Teacher Model.} The goal of the teacher is to provide \textit{soft} pseudo-labels on unlabeled sets of video-text candidate pairs. We adopt CLIP~\cite{clip} as a teacher. CLIP includes an image encoder and a text encoder, which were trained to predict the correct pairing of image-text pairs using a contrastive objective \cite{infonce}. In practice, we use CLIP to compute the similarity between a subset of videos (within a large unlabeled set) and a set of candidate texts. Given that CLIP only takes individual images as input, we pass $N$ frames from the video through its visual encoder and mean-pool the outputs into a single visual feature. We then use these similarity scores as target soft pseudo-labels.

\noindent\textbf{Student Model.} We aim to train a student model that learns from video-text pairs and distills knowledge from large pretrained image-language models. As the student, we choose the same dual architecture proposed by CLIP~\cite{clip}. To train the model, we leverage two types of supervision: samples from the manually labeled video-text pairs dataset and soft pseudo-labels from the unlabeled set. Like the Teacher model, the student's visual stream takes $N$ frames from each video and mean-pool the resulting representations into a single feature.

\noindent\textbf{Student's Training Objective.} We train the student model with two losses: a loss to learn from labeled samples, and a loss to distill the teacher knowledge via pseudo-labels. Given a video-text pair denoted $(v,t)$ our student's dual encoder extracts a video representation $z_v$ and a text representation $z_t$. For labeled samples, we use the InfoNCE~\cite{infonce} loss to learn a video-text correspondence. We follow \cite{frozen_in_time,videoclip} and minimize the text-to-video and video-to-text contrastive losses:

\begin{equation}
\label{eq:v2t}
    \mathcal{L}_{v2t} = \sum_{(v, t)\in B_l}\log{\frac{e^{z_v \cdot z_t^+ / \sigma}}{\sum_{z \in \{z_t^+, z_t^-\}} e^{z_v \cdot z / \sigma}}}
\end{equation}

\begin{equation}
\label{eq:t2v}
    \mathcal{L}_{t2v} = \sum_{(v, t)\in B_l}\log{\frac{e^{z_t \cdot z_v^+ / \sigma}}{\sum_{z \in \{z_v^+, z_v^-\}} e^{z_t \cdot z / \sigma}}}
\end{equation}

where $\sigma$ is the temperature hyper-parameter, $B_l$ is a batch of video-text pairs, $z_t^+$ the positive text for the candidate video $z_v$, $z_v^+$ the positive video for candidate text $z_t$, and $\{z_{v}^{-}, z_{t}^{-}\}$ the negatives sets to contrast the candidate video and text representations. Then $(\mathcal{L}_{v2t} + \mathcal{L}_{t2v})$ is the final labeled (contrastive) loss.

To distill knowledge from soft pseudo-labels generated by the teacher, we use the teacher's predictions as pseudo-labels \cite{distillation} and minimize the cross-entry of the student's scores relative to those from the teacher: %

\begin{equation}
\small
\label{eq:distill-a}
    \mathcal{L}_{distill,v2t} = \sum_{(v, t)\in B_l} \frac{e^{x_v \cdot x_t / \sigma}}{\sum_{x \in T} e^{x_v \cdot x / \sigma}} \log{\frac{e^{z_v \cdot z_t / \sigma}}{\sum_{z \in T} e^{z_v \cdot z / \sigma}}}
\end{equation}

\begin{equation}
\small
\label{eq:distill-b}
    \mathcal{L}_{distill,t2v} = \sum_{(v, t)\in B_l} \frac{e^{x_v \cdot x_t / \sigma}}{\sum_{x \in V} e^{x \cdot x_t / \sigma}} \log{\frac{e^{z_v \cdot z_t / \sigma}}{\sum_{z \in V} e^{z \cdot z_t / \sigma}}}
\end{equation}

where $x_v$ and $x_t$ are the teacher's video and text representations, and $V$ and $T$ are the sets of videos and texts in the batch.

Our final objective combines the contrastive and distillation losses as in \cref{eq:objective}. We scale the distillation loss, with $\lambda$, to prevent over-fitting to noisy pseudo-labels.

\begin{equation}
\label{eq:objective}
    \mathcal{L} = \lambda (\mathcal{L}_{distill,v2t} + \mathcal{L}_{distill,t2v}) + (1 - \lambda) (\mathcal{L}_{v2t} + \mathcal{L}_{t2v})
\end{equation}

\subsection{Fusing Teacher-Student Knowledge}

We aimed to train a competent student compared to the teacher. However, it is hard to compete with the 400M image-text pairs that were used to train CLIP~\cite{clip}. Therefore, our goal is, instead, to fuse both: the general visual knowledge encapsulated by the teacher and the video-specific properties learned by the student. There are multiple ways to ensemble models \cite{dietterich2000ensemble}; however, given that our fine-tuning strategy gently adapts the teacher to video use cases, we can leverage elegant weight-space ensembling techniques \cite{garipov2018loss,wortsman2021robust}. We follow the same approach in~\cite{wortsman2021robust} to linearly combine the teacher and student weights (by $\alpha$) and create our final model, FitCLIP.

\subsection{FitCLIP's Implementation Details}

We uniformly sample $N=4$ frames from each video, similarly to TSN~\cite{wang2016temporal}. The Teacher and Student models both use a VIT-B/16 architecture initialized with OpenAI's publicly released weights \cite{clip}. We empirically set $\lambda={10}^{-4}$ to smooth the training process (note the labeled and pseudo-labeled loss magnitudes may be wildly different). We consistently use $\sigma=0.05$ as temperature value. At training time, we randomly crop the frames to a size of $224\times224$, and perform random horizontal flips. We use the AdamW optimizer with a learning rate equal to $3\times10^{-5}$. We use the same tokenizer as in CLIP~\cite{clip}. We conduct our experiments using 8x A100 (40GB) GPUs. We use 4.5K labeled videos, randomly sampled from the WebVid-2.5M dataset~\cite{frozen_in_time}, to compute the losses in \cref{eq:v2t,eq:t2v}. The entire WebVid-2.5M dataset is used to compute the distillation losses -- \cref{eq:distill-a,eq:distill-b}. We choose the (labeled) validation loss in the WebVid-2.5M dataset as a criterion to select the best student models. Finally, to fuse the teacher and the student weights, we use $\alpha=0.4$. We encourage the reader to read the supplementary material for analyses to some of the hyperparameter values. We wrote our code on Python using PyTorch~\cite{pytorch} and Lightning~\cite{lightning}.

\nocite{hf-transformers,hydra,matplotlib,nltk,pandas,parallel,seaborn,spacy}

\section{Zero-shot Video Understanding Benchmark}%
\label{sec:benchmark}

\subsection{Baselines}%
\label{sec:baselines}

\noindent\textbf{CLIP~\cite{clip}.} This model has been pre-trained with the WIT dataset~\cite{wit}, which contains about 400M image-text pairs. We re-implement the zero-shot inference of this baseline model. To deal with video, we encode $N=4$ uniformly sampled frames per video and average their features to obtain the final video representation. In all our experiments, we use the publicly released CLIP ViT-B/16~\cite{vit} model. Note that our CLIP adaptation is equivalent to ActionCLIP ~\cite{actionclip} (see \textit{Supplementary Material}).

\noindent\textbf{CLIP4Clip~\cite{clip4clip}.} This method proposes changes on top of CLIP. In particular, they propose something the authors call \emph{post-pretraining} that fine-tunes CLIP on the category ``Food and Entertaining'' (380k videos) from the HowTo100M~\cite{howto100m} dataset. The authors have not provided this checkpoint, so we cannot evaluate it on our benchmarks. Still, we decide to include the results they report. Nevertheless, note the evaluation conditions are not the same to constitute a fair comparison (\eg{}, the authors sample more than 4 frames per video clip).

\noindent\textbf{Frozen in Time~\cite{frozen_in_time} (Frozen).} This model was pre-trained leveraging video-text pairs from the WebVid dataset. There are multiple versions pre-trained versions for this model, including one that leverages the well-curated CC3M image-text pairs dataset. In our (main) experiments, we use the model that trains using the WebVid-2.5M, COCO, and CC3M dataset (note this is much less data than CLIP's pretraining dataset). Results for other versions of Frozen in Time can be found in the supplementary material.

\noindent\textbf{VideoCLIP~\cite{videoclip}.} This baseline uses a Transformer~\cite{transformer} on top of a frozen HowTo100M-pre-trained S3D~\cite{s3d} video model from MIL-NCE~\cite{mil_nce} and a fine-tuned BERT~\cite{bert} text model. This method trains on HowTo100M. A notable difference is that VideoCLIP samples 32 clips of size 32 frames (1024 frames) for each video, while we sample only 4 frames for each video.

\noindent\textbf{VIOLET~\cite{violet}.} This method uses a video-language transformer trained end-to-end by masking discrete visual tokens. The authors use multiple training datasets including CC3M and WebVid.

\noindent\textbf{BridgeFormer~\cite{bridgeformer} (BF).} This model leverages a multimodal encoder on top of the unimodal encoders and a method that masks the main verb and nouns as a for of multiple-choice questions as a pre-text task. The authors find this method to be more sample efficient than vanilla NCE.

\subsection{Zero-shot Tasks and Datasets}

\noindent\textbf{Action Recognition.} Our goal is to classify a video with one of $C$ possible action classes. To do so, we form pretext language queries with predefined prompts. An illustrative example is the following prompt: "a video of a person $\{c_i\}$", where $c_i$ is the $i$-th class out of the $C$ candidate action categories. Given the visual representation of the target video, we compute its similarity with the language feature of each candidate action class prompt. We predict the action class by selecting the visual-text pair with the highest similarity.
We report the top-1 and top-5 accuracy.
We evaluate zero-shot action recognition in two datasets:
\begin{itemize}
    \item \emph{Moments in Time (MiT)~\cite{moments_in_time}} consists of 3-second YouTube clips that capture the dynamics of actions performed by varied subjects including animals and humans. The dataset includes $339$ categories and $33,900$ validation videos.
    \item \emph{UCF101~\cite{ucf101}} contains 101 action classes. Our zero-shot experiments in this dataset aim to classify all the $1794$ available test videos from the split 1.
\end{itemize}

\noindent\textbf{Text-to-video Retrieval.} Given a text query, the goal of text-to-video retrieval is to find a video, from a collection, that visually matches the text description. Given that the concept of classes does not exist in this task, previous methods \cite{frozen_in_time,clip4clip} denote experiments as zero-shot when the visual-language models are not fine-tuned on the downstream datasets. To measure performance, we report recall at $k=\{1,5,10\}$ and the median ranking (MdR). We evaluate zero-shot text-to-video retrieval in three datasets:
\begin{itemize}
    \item \emph{MSR-VTT~\cite{msrvtt}} contains video clips with a duration of up to 30 seconds paired with captions. We adopt the 1K-A test split~\cite{jsfusion}, which contains $1,000$ video-text pairs.
    \item \emph{YouCook2~\cite{youcook2}}  comprises challenging cooking videos depicting fine-grained human actions. We test on $3305$ clip-text pairs~\cite{mil_nce}.
    \item \emph{DiDeMo~\cite{didemo}} contains mostly unedited video clips from Flickr. We follow \cite{collaborative_experts,clipbert,frozen_in_time} and cast a video-paragraph retrieval problem. We evaluate on $4021$ test samples.
\end{itemize}  

\section{Experimental Results}%
\label{sec:experiments}

In this section, we conduct zero-shot experiments in two popular video understanding tasks, and then a diagnostic analysis of FitCLIP. First, we study the performance of the zero-shot baselines described in \cref{sec:baselines} in the task of action recognition. The second analysis summarizes the baseline performance in diverse datasets for text-to-video retrieval. We run diagnostic experiments to validate the importance of fusing the teacher knowledge, as in \cite{wortsman2021robust}, to a competent zero-shot model. Finally, we run performance analyses on FitCLIP that study per-class gains in the action recognition task, and the shift in ranking distributions for the text-to-video retrieval tasks.

\subsection{Zero-shot Action Recognition Results}

We compare the zero-shot performance of FitCLIP and different baselines using two popular action recognition datasets. We describe the results and provide our analysis.

\noindent\textbf{Analysis on Moments in Time.} \Cref{tab:mit-results} summarizes the zero-shot results in the moments in time dataset. To establish a reference point, we also report VATT~\cite{vatt}, the state-of-the-art using full supervision. In this dataset, FitCLIP outperforms both baselines, CLIP and Frozen, by a significant margin. It is noteworthy that CLIP, without seeing video data at training time, still outperforms Frozen by $11\%$ at top-5 accuracy. Despite CLIP's good performance, we observe that FitCLIP further improves performance by $4.3\%$ (top-5) setting a new state-of-the-art in this dataset. While FitCLIP achieves outstanding zero-shot results, a \eg{} $44.6\%$ top-5 accuracy, there is still an ample gap with respect to approaches that leverage supervision from the target dataset.

\noindent\textbf{Analysis on UCF101.} \Cref{tab:ucf101-results} shows the results on the UCF101 zero-shot benchmark. FitCLIP outperforms CLIP at Top 5 accuracy and sligthly underperforms at Top 1. All the findings remain consistent: a not-so-large gap between the best zero-shot and supervised approaches and Frozen under-performing with respect to CLIP-based methods. We attribute FitCLIP and CLIP close performance (when looking at both top-1 and top-5) due to the characteristics of UCF101, which contains a lot of common actions, including many sport-related actions. These types of actions often appear in photographs, and chances are, they are well-represented in CLIP training set.

\begin{table}[ht!]
\centering
\subfloat[
\textbf{Moments in Time (MiT)}
\label{tab:mit-results}
]{
\centering
\begin{minipage}{0.45\columnwidth}{\begin{center}
\tablestyle{6pt}{1.0}
\begin{tabular}{x{36}z{38}z{38}}
Method & Top 1 & Top 5 \\
\shline
\multicolumn{3}{l}{\gc{Supervised}} \\
VATT~\cite{vatt} & 41.1 & 67.7 \\
\hline
\multicolumn{3}{l}{\gc{Zero-shot}} \\
Frozen & 14.0 & 31.8 \\
CLIP & 19.9 & 40.3 \\
FitCLIP & \textbf{21.8} & \textbf{44.6} \\
\end{tabular}
\end{center}}\end{minipage}
}
\subfloat[
\textbf{UCF101}
\label{tab:ucf101-results}
]{
\centering
\begin{minipage}{0.45\columnwidth}{\begin{center}
\tablestyle{6pt}{1.0}
\begin{tabular}{x{48}z{36}z{36}}
Method & Top 1 & Top 5 \\
\shline
\multicolumn{3}{l}{\gc{Supervised}} \\
SMART~\cite{Gowda_Rohrbach_Sevilla-Lara_2021} & 98.6 & -- \\
\hline
\multicolumn{2}{l}{\gc{Zero-shot}} \\
Frozen & 51.9 & 76.1 \\
BF~\cite{bridgeformer} & 51.1 & -- \\
CLIP & \textbf{74.5} & 94.3 \\
FitCLIP & 73.3 & \textbf{95.3} \\
\end{tabular}
\end{center}}\end{minipage}
}

\caption{\textbf{Zero-shot action recognition results.} (a) FitCLIP improves performance upon CLIP, and significantly outperforms Frozen. 
(b) FitCLIP shows slight improvements upon CLIP; Frozen lags behind in terms of zero-shot performance. Reported numbers in both tables are percentages and compute the top-1 and top-5 accuracy.
}
\label{tab:action-recognition-results-mit}
\end{table}

\subsection{Zero-shot Text-to-video Retrieval}

\begin{table}
\centering
\subfloat[
\textbf{\scriptsize MSR-VTT}
\label{tab:msrvtt-results}
]{
\centering

\resizebox{.31\textwidth}{!}{%
\tablestyle{4pt}{1.0}
\begin{tabular}{x{53}z{18}z{18}z{18}z{18}}
Method & R@1 & R@5 & R@10 & MdR \\
\shline
\multicolumn{2}{l}{\gc{Supervised}}\\
CAMoE~\cite{camoe} & 52.9 & 78.5 & 86.5 & 1\\
\hline
\multicolumn{2}{l}{\gc{Zero-shot}}\\
VideoCLIP~\cite{videoclip} & 10.4 & 22.2 & 30.0 & -- \\
Frozen & 21.3 & 43.6 & 55.9 & 7 \\
VIOLET~\cite{violet} & 25.9 & 49.5 & 59.7 & -- \\
BF~\cite{bridgeformer} & 26.0 & 46.4 & 56.4 & 7 \\
CLIP via~\cite{clip4clip} & 30.6 & 54.4 & 64.3 & 4 \\
CLIP4Clip~\cite{clip4clip} & 32.0 & 57.0 & 66.9 & 4 \\
CLIP & 30.4 & 55.1 & 64.1 & 4 \\
FitCLIP & \textbf{33.8} & \textbf{59.8} & \textbf{69.4} & \textbf{3} \\
\end{tabular}
}
}
\subfloat[
\textbf{\scriptsize YouCook2}
\label{tab:ycook-results}
]{
\centering
\resizebox{.31\textwidth}{!}{%
\tablestyle{4pt}{1.0}
\begin{tabular}{x{42}z{18}z{18}z{18}z{18}}
Method & R@1 & R@5 & R@10 & MdR \\
\shline
\multicolumn{2}{l}{\gc{Supervised}}\\
TACo~\cite{taco} & 29.6 & 59.7 & 72.7 & 4 \\
\hline
\multicolumn{2}{l}{\gc{Zero-shot}}\\
VideoCLIP~\cite{videoclip} & \textbf{22.7} & \textbf{50.4} & \textbf{63.1} & -- \\
Frozen & 3.2 & 10.1 & 16.2 & 135 \\
CLIP & 5.3 & 14.6 & 20.9 & 94 \\
FitCLIP & 5.8 & 15.5 & 22.1 & 75 \\
\end{tabular}
}
}
\subfloat[
\textbf{\scriptsize DiDeMo}
\label{tab:didemo-results}
]{
\centering
\resizebox{.31\textwidth}{!}{%
\tablestyle{4pt}{1.0}
\begin{tabular}{x{42}z{18}z{18}z{18}z{18}}
Method & R@1 & R@5 & R@10 & MdR \\
\shline
\multicolumn{2}{l}{\gc{Supervised}}\\
CAMoE~\cite{camoe} & 43.8 & 71.4 & 79.9 & 2\\
\hline
\multicolumn{2}{l}{\gc{Zero-shot}}\\
VideoCLIP~\cite{videoclip} & 16.6 & 46.9 & -- & -- \\
Frozen & 23.2 & 45.8 & 56.8 & 7 \\
VIOLET~\cite{violet} & 23.5 & 49.8 & 59.8 & -- \\
BF~\cite{bridgeformer} & 25.6 & 50.6 & 61.1 & 5 \\
CLIP & 26.2 & 49.9 & 60.6 & 5 \\
FitCLIP & \textbf{28.5} & \textbf{53.7} & \textbf{64.0} & \textbf{4} \\
\end{tabular}
}
}
\caption{\textbf{Zero-shot text-to-video retrieval results.} In all datasets, FitCLIP improves upon CLIP by significant margins. (a) FitCLIP's shows the best zero-shot results though there is an important gap with the supervised state of the art.
(b) In this dataset, YouCook2, FitCLIP exhibits the largest gap concerning fully supervised approaches and with VideoCLIP, that pretrained on HowTo100M. We attribute this result to the fine-grained nature of the dataset. 
(c) FitCLIP shows consistent boosts upon CLIP even for the DiDeMo (paragraph-retrieval) task, which includes long language queries.
R@k denotes recall at the top-$k=\{1,5,10\}$ predictions, and MdR refers to the Median Ranking metric.}
\label{tab:action-recognition-results-didemo}
\end{table}

To compare FitCLIP and the baselines, here we report the experimental results and analysis for the text-to-video retrieval task.

\noindent\textbf{Analysis on MSR-VTT.} \Cref{tab:msrvtt-results} summarizes results in the MSR-VTT dataset. We observe that Frozen performance is poor compared to that of CLIP and FitCLIP. Even though Frozen was trained on video data with similar properties to MSR-VTT, it is hard for this model to compete with the general knowledge encoded in CLIP-like models. We observe FitCLIP consistently improves performance upon CLIP across all the retrieval metrics. These results suggest that FitCLIP captures complementary video-language information that CLIP lacks. Concerning the gap to reach the performance of the best-supervised approach, CAMoE \cite{camoe}, FitCLIP does not lag that behind. Even though there is a $16.1\%$ gap at R@10, we see that FitCLIP closely approaches supervised performance at the MdR metric.

\noindent\textbf{Analysis on YouCook2.} We report zero-shot results for the YouCook2 dataset in \cref{tab:ycook-results}. From the get-go, we observe the difficulty of this dataset. Even the state-of-the-art, TACo~\cite{taco}, struggles to achieve more than $30\%$ R@1. While we observe that FitCLIP's performance consistently outperforms other zero-shot baselines, we have observed a large overall gap between our method and those that are supervised or pretrained on HowTo100M~\cite{howto100m} (VideoCLIP~\cite{videoclip} in the table). We hypothesize this is due to the fine-grained nature of the language descriptions contained in YouCook2 and HowTo100M. Moreover, lots of videos in this dataset are captured from an egocentric view.

\noindent\textbf{Analysis on DiDeMo.} \Cref{tab:didemo-results} summarizes the results in DiDeMo's paragraph retrieval task. First, we observe that the performance of Frozen, VIOLET~\cite{violet}, and BridgeFormer~\cite{bridgeformer} approach the one achieved by CLIP in this dataset. Contrary to other datasets, DiDeMo contains unedited, human-centric footage that shares commonalities with the WebVid dataset used to train Frozen. Conversely, FitCLIP, which leverages both: the knowledge from CLIP and the WebVid dataset, achieves the best performance overall. For completeness, we report the CAMoE's supervised performance \cite{camoe}, which is $15.9\%$ better than FitCLIP, the most competitive zero-shot alternative.

The results on these three datasets empirically demonstrate the value of FitCLIP to push the limits of zero-shot text-to-video retrieval. FitCLIP establishes a new state-of-the-art in zero-shot text-to-video retrieval across three different datasets. Despite such a milestone, there is still room for improvement, especially in fine-grained datasets such as YouCook2. We hope this benchmark promotes more work on zero-shot text-to-video retrieval.

\subsection{Diagnostic Analysis}

\noindent\textbf{Impact of Fusing the Teacher-Student Knowledge (\cref{tab:teacher-student}).} One of the key properties of FitCLIP is the ability to incorporate the student learning from video data, and knowledge of the CLIP teacher. Here we report the performance of both our Student and Teacher (CLIP) and contrast that with the final zero-shot performance obtained with FitCLIP. \Cref{tab:teacher-student} summarizes the results. We observe that although the Student's performance remains inferior to that of the Teacher, it is close enough in various datasets, \eg{} MiT, MSR-VTT, and DiDeMo. $\triangle$ denotes the difference in performance between FitCLIP and the teacher and indirectly measures the contribution of the student learning. We observe that improvements are consistent across all tasks and datasets. These results suggest that the Student effectively passes complementary information to the teacher after weight assembling.

\noindent\textbf{Additional Ablations.} Due to space limitations, we include additional analysis in the \textit{Supplementary Material}. We compare the properties of FitCLIP vs. CLIP, do a deep-dive on the impact of fusing the Teacher-Student knowledge, ablate weight-ensembling parameters, and report comparisons with additional methods trained on HowTo100M.

\begin{table}
\centering
\tablestyle{3.5pt}{1.05}
\begin{tabular}{c|rr|rrr}
& \multicolumn{2}{c|}{Action Recognition} & \multicolumn{3}{c}{Text-to-video Retrieval}  \\
& UCF101 & MiT & MSR-VTT & YouCook2 & DiDeMo\\
\shline
Teacher (CLIP) & 74.5 & 19.9 & 55.1 & 14.6 & 49.9 \\
Student & 64.7 & 17.7 & 52.6 & 9.7 & 42.4 \\
\hline
FitCLIP & 73.3 & 21.8 & 59.8 & 15.5 & 53.7 \\
$\triangle$ & \textcolor{red}{\(\downarrow\) 1.2} & \textcolor{nice-green}{\(\uparrow\) 1.9} & \textcolor{nice-green}{\(\uparrow\) 4.7} & \textcolor{nice-green}{\(\uparrow\) 0.9} & \textcolor{nice-green}{\(\uparrow\) 3.8} \\
Err\onedot rate red\onedot & \textcolor{red}{\(\downarrow\) 4.7} & \textcolor{nice-green}{\(\uparrow\) 2.4} & \textcolor{nice-green}{\(\uparrow\) 10.5} & \textcolor{nice-green}{\(\uparrow\) 1.1} & \textcolor{nice-green}{\(\uparrow\) 7.6} \\
\end{tabular}
\caption{\textbf{Impact of fusing teacher-student knowledge.} $\triangle$ denotes the absolute difference in performance between FitCLIP and the Teacher model. We report the top-1 accuracy for the zero-shot action recognition datasets, and the top-5 recall for the zero-shot text-to-video retrieval ones. We observe that even though the Student model is weaker than the Teacher, it still provides complementary information to FitCLIP, yielding consistent improvements ($\triangle$) across datasets. Full results with all the metrics in the supplementary material.}
\label{tab:teacher-student}
\end{table}

\section{Conclusions}

This paper presents a fine-tuning strategy to adapt large-scale image-text pre-trained models for zero-shot video understanding tasks, dubbed FitCLIP. FitCLIP performs well on zero-shot settings for three Text-to-Video Retrieval and two Action Recognition tasks that we evaluated. We show the importance of doing the weight-space ensembling step of our method to keep or improve the teacher's robust performance across different datasets, even when the student was trained on different data.
We highlight our method introduces no extra inference costs while improving CLIP results overall.

\section*{Acknowledgements}

We thank Christine Feak for revising a draft of this document. Santiago thanks Adobe Research for providing financial support to continue working on this project after the internship finished.

\bibliography{egbib}
\end{document}


\maketitle

\renewcommand*{\thefootnote}{\fnsymbol{footnote}}
\footnotetext[1]{Work done as an intern at Adobe Research.}
\renewcommand*{\thefootnote}{\arabic{footnote}}
\setcounter{footnote}{0}

\section{Pretraining Datasets}

\Cref{tab:pretrain-dataset} summarizes existing datasets for pretraining visual-language models. CC3M~\cite{cc3m} is one of the first datasets to bridge images with natural language supervision leveraging the internet (HTML image alt texts). This dataset collects about 3M clean images through a pipeline that warranty clean supervision signal. The MS COCO Captions~\cite{coco-captions} (COCO) dataset contains ~500k human-curated caption-image pairs. The images come from the MS COCO~\cite{coco} dataset, which in turn were collected from Flickr. WIT~\cite{wit} contains 37.5M image-caption pairs obtained from the Wikipedia. CLIP authors~\cite{clip} constructed dataset that contains more than 400M text-image pairs scrapped from the internet. The dataset contains images retrieved from queries formed with the $1000$ most common visual concepts in Wikipedia. While the dataset does not rely on manual cleaning to verify the image-text pairs, it is assumed that a person provided a good enough image caption before uploading the image to the internet. In the same spirit, the WebVid-2.5M dataset \cite{frozen_in_time} crawls 2.5M text-video pairs leveraging manually-curated titles from Stock footage. Differently, the HowTo100M (HT100M) dataset \cite{howto100m} contains 100M pairs of noisy aligned video-text pairs. In this dataset, the video-text pairs come from long YouTube videos and their automatically transcribed speech.

\begin{table}
\centering
\subfloat[
\textbf{\scriptsize Pretraining datasets}
\label{tab:pretrain-dataset}
]{
\centering
\resizebox{.3\textwidth}{!}{%
\tablestyle{4pt}{1.05}
\begin{tabular}{x{48}x{28}x{35}z{32}}
Dataset & Domain & Supervision & Size \\
\shline
COCO~\cite{coco-captions} & Images & Clean & 600k \\
CC3M~\cite{cc3m} & Images & Clean & 3M \\
WIT~\cite{wit} & Images & Clean & 37.5M \\
CLIP~\cite{clip} & Images & Weak & 400M \\
WebVid~\cite{frozen_in_time} & Videos & Weak & 2.5M \\
HT100M~\cite{howto100m} & Videos & Noisy & 100M \\
\end{tabular}
}
}
\subfloat[
\textbf{\scriptsize ZS action recognition}
\label{tab:action-recognition}
]{
\centering
\resizebox{.3\textwidth}{!}{%
\tablestyle{4pt}{1.05}
\begin{tabular}{x{48}z{45}z{45}}
Dataset & \# Classes & \# Samples \\
\shline
MiT~\cite{moments_in_time} & 339 & 33,900 \\
UCF101~\cite{ucf101} & 101 & 1,794 \\
\multicolumn{3}{c}{~}\\
\multicolumn{3}{c}{~}\\
\end{tabular}
}
}
\hspace{1pt}
\subfloat[
\textbf{\scriptsize ZS text-to-video retrieval}
\label{tab:retrieval}
]{
\centering
\resizebox{.32\textwidth}{!}{%
\tablestyle{4pt}{1.05}
\begin{tabular}{x{62}z{45}x{36}}
Dataset & \# Samples & Genre \\
\shline
MSR-VTT~\cite{msrvtt} & 1000 & UGC \\
YouCook2~\cite{youcook2} & 3305 & Cooking \\
DiDeMo~\cite{didemo} & 4021 & UGC \\
\multicolumn{3}{c}{~}\\
\end{tabular}
}
}
\caption{\textbf{Pretraining and Zero-shot Datasets.} (a) Diverse image and video datasets are available for pretraining visual-language models.
(b) We benchmark zero-shot (ZS) action recognition in two popular datasets. MiT denotes Moments in Time \cite{moments_in_time}. 
(c) To benchmark zero-shot (ZS) text-to-video retrieval, we rely on three well-established datasets. UGC stands for user-generated content, and Genre refers to the type of videos in the dataset.}
\label{tab:datasets}%
\end{table}

\section{FitCLIP \vs{} CLIP per-class performance}

\begin{figure}
\centering

\subfloat[
\textbf{Bottom-25 Classes}
\label{tab:diff-mit1}
]{
\centering
\includegraphics[width=.45\textwidth]{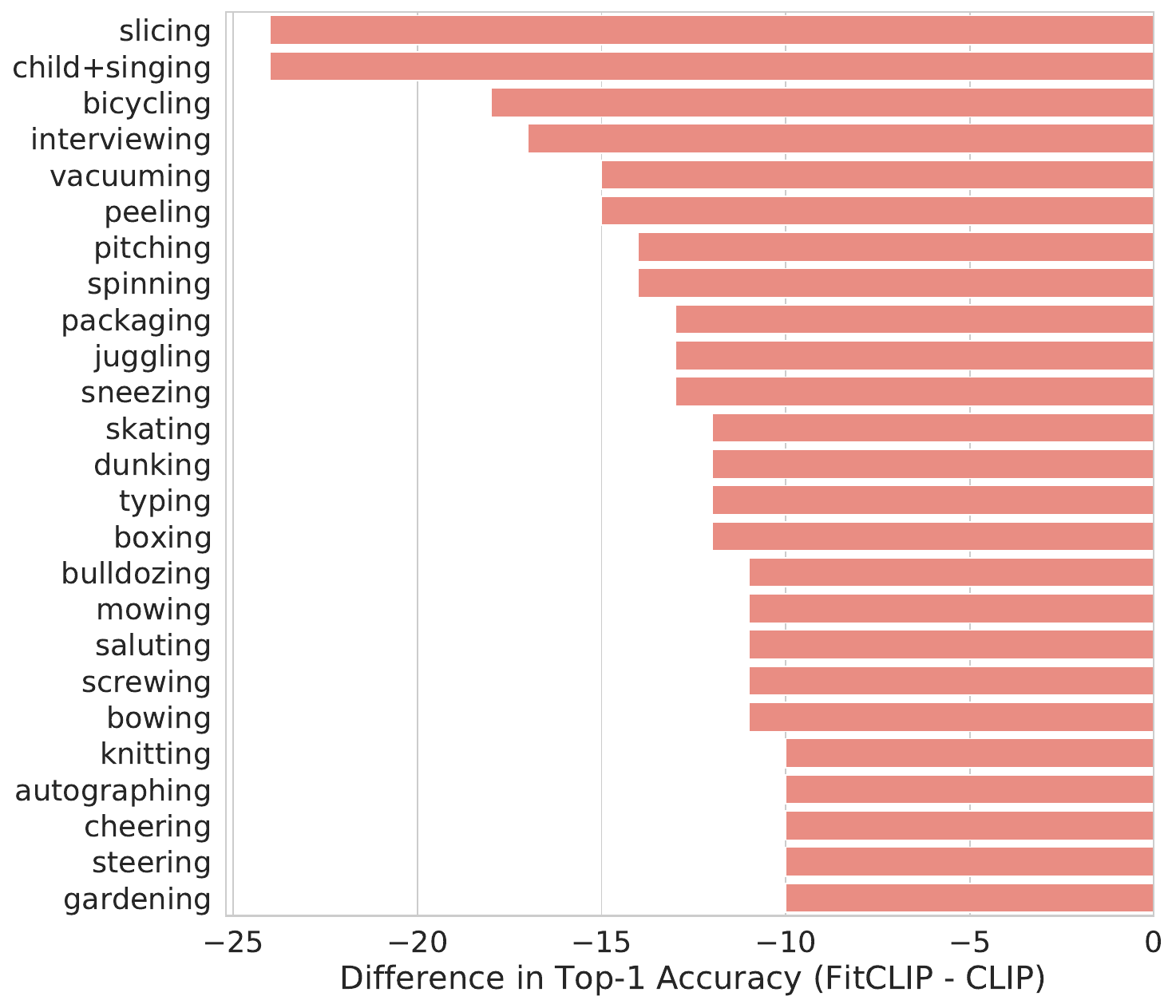}
}
\subfloat[
\textbf{Top-25 Classes}
\label{tab:diff-mit2}
]{
\centering
\includegraphics[width=.45\textwidth]{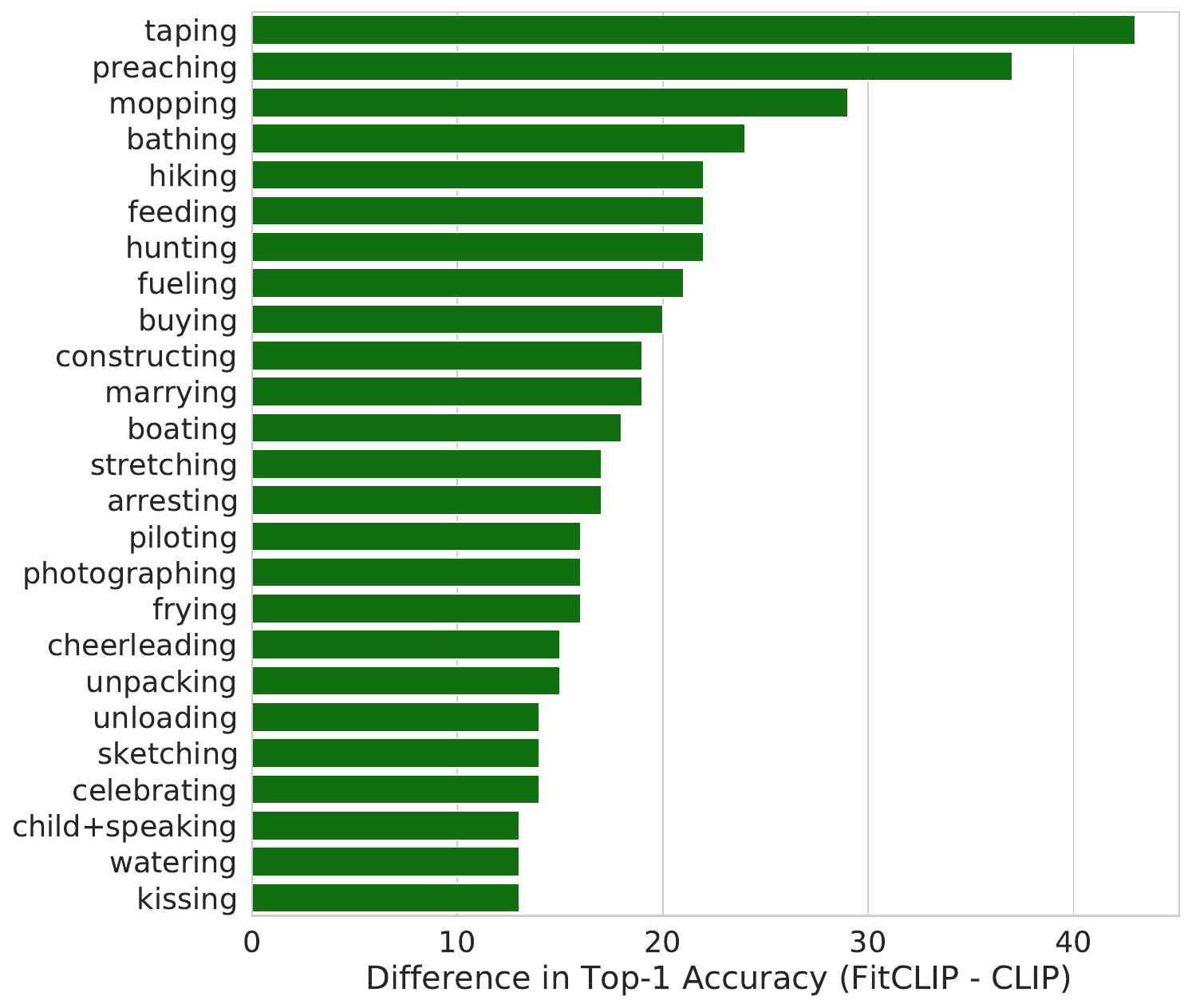}
}

\caption{\textbf{FitCLIP \textit{vs.} Teacher per-class improvements.} The plots show the per-class difference between FitCLIP and CLIP performances (Top-1) on the Moments in Time (MiT) dataset. Noticeably, the performance difference varies significantly across various action classes, which reinforce our intuition that FitCLIP encodes complementary video information compared to CLIP. Interestingly, FitCLIP improves performance for abstract action classes such as \textit{preaching and tapping}, while CLIP does so for actions involving common actions like \textit{cycling, boxing, or skating}.} 
\label{fig:diff-mit}%
\end{figure}

Previous experiments showed that FitCLIP offers a simple strategy to boost zero-shot performance in video understanding tasks; however, where are those improvements emerging from? To understand better the differences between FitCLIP and CLIP (which is also our teacher), we compute the performance difference per class, between both models, in the Moments in Time dataset. \Cref{fig:diff-mit} summarizes the results by plotting the largest and smallest (including actions with worst performance) 25 changes in performance. First, we observe that the performance accuracy (Top-1) of several classes changes drastically. This validates our hypothesis that the Student provides FitCLIP with complementary information concerning the knowledge CLIP (the Teacher) already provides. Interestingly, FitCLIP obtains, overall, better performance for abstract action classes such as \textit{preaching and taping}. On the contrary, CLIP tends to do better for common actions often captured in photographs such as \textit{skating, or boxing}.

\section{FitCLIP \vs{} CLIP ranking distributions}

\begin{figure}
\begin{center}
\includegraphics[width=.8\textwidth]{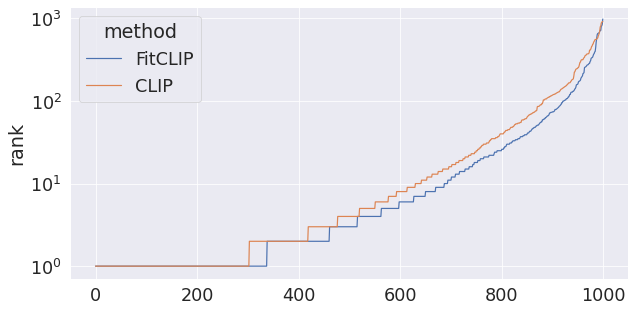}
\end{center}
\caption{\textbf{FitCLIP vs. CLIP distribution of Text-to-Video Retrieval rankings.} The x-axis represents each text in the MSR-VTT validation set (1K-A split) and the y-axis (in log scale) represents the rank each model gave to the corresponding video. The x-axis is sorted by rank (the values increase).}
\label{fig:rank}
\end{figure}

The Text-to-Video Retrieval results show that FitCLIP outperforms CLIP at multiple points of this zero-shot setting. However, it is not clear how the methods behave for the rest of them. \Cref{fig:rank} shows the distribution of rankings for the validation set of MSR-VTT for both methods. We can see that FitCLIP is under the CLIP curve for virtually all points. FitCLIP ranks the videos better for this dataset, regardless of the cutting point.

\section{Frozen in Time Variants}

\Cref{tab:action-recognition-frozen,tab:zero-shot-results-frozen} show the results on zero-shot action recognition and text-to-video retrieval for Frozen in Time~\cite{frozen_in_time} on different pre-training datasets. These pre-trained checkpoints are provided by the authors\footnote{\url{https://github.com/m-bain/frozen-in-time\#-pretrained-weights}}. They use different combinations of Conceptual Captions~\cite{cc3m} (CC3M), WebVid~\cite{frozen_in_time}, and Microsoft COCO Captions~\cite{coco-captions} (COCO). Combining the three of them presents the best results. However, note COCO Captions were obtained using an expensive data collection procedure and are richly annotated while the other two datasets were obtained from data available on the internet and thus have weaker annotations.

\begin{table}
\centering
\subfloat[
\textbf{Moments in Time (MiT)}
\label{tab:mit-results-frozen}
]{
\centering
\begin{minipage}{0.9\columnwidth}{\begin{center}
\tablestyle{6pt}{1.0}
\begin{tabular}{crr}
Dataset & Top 1 & Top 5 \\
\shline
WebVid & 11.4 & 27.2 \\
CC3M+WebVid & 13.2 & 29.3 \\
CC3M+WebVid+COCO & \textbf{14.0} & \textbf{31.8} \\
\end{tabular}
\end{center}}\end{minipage}
}
\vspace{1em}
\subfloat[
\textbf{UCF101}
\label{tab:ucf101-results-frozen}
]{
\centering
\begin{minipage}{0.9\columnwidth}{\begin{center}
\tablestyle{6pt}{1.0}
\begin{tabular}{crr}
Dataset & Top 1 & Top 5 \\
\shline
WebVid & 36.9 & 61.1 \\
CC3M+WebVid & 49.2 & 61.1 \\
CC3M+WebVid+COCO & \textbf{51.9} & \textbf{76.1} \\
\end{tabular}
\end{center}}\end{minipage}
}
\caption{Zero-shot action recognition results of Frozen in Time~\cite{frozen_in_time} pre-trained on different datasets.
}
\label{tab:action-recognition-frozen}
\end{table}

\begin{table}
\centering
\subfloat[
\textbf{MSR-VTT}
\label{tab:msrvtt-results-frozen}
]{
\centering
\begin{minipage}{0.9\linewidth}{\begin{center}
\tablestyle{6pt}{1.0}
\begin{tabular}{crrrr}
Dataset & R@1 & R@5 & R@10 & MdR \\
\shline
WebVid & 12.9 & 31.0 & 41.2 & 16 \\
CC3M+WebVid & 17.1 & 39.1 & 49.6 & 11 \\
CC3M+WebVid+COCO & \textbf{21.3} & \textbf{43.6} & \textbf{55.9} & \textbf{7} \\
\end{tabular}
\end{center}}\end{minipage}
}
\vspace{1em}
\subfloat[
\textbf{YouCook2}
\label{tab:ycook-results-frozen}
]{
\centering
\begin{minipage}{0.9\linewidth}{\begin{center}
\tablestyle{6pt}{1.0}
\begin{tabular}{crrrr}
Dataset & R@1 & R@5 & R@10 & MdR \\
\shline
WebVid & 1.1 & 4.2 & 6.8 & 329 \\
CC3M+WebVid & 2.7 & 9.5 & 14.2 & 162 \\
CC3M+WebVid+COCO & \textbf{3.2} & \textbf{10.1} & \textbf{16.2} & \textbf{135} \\
\end{tabular}
\end{center}}\end{minipage}
}
\vspace{1em}
\subfloat[
\textbf{DiDeMo}
\label{tab:didemo-results-frozen}
]{
\centering
\begin{minipage}{0.9\linewidth}{\begin{center}
\tablestyle{6pt}{1.0}
\begin{tabular}{crrrr}
Dataset & R@1 & R@5 & R@10 & MdR \\
\shline
WebVid & 14.5 & 34.9 & 45.4 & 14 \\
CC3M+WebVid & 20.3 & 42.7 & 53.5 & 9 \\
CC3M+WebVid+COCO & \textbf{23.2} & \textbf{45.8} & \textbf{56.8} & \textbf{7} \\
\end{tabular}
\end{center}}\end{minipage}
}
\caption{Zero-shot text-to-video retrieval results of Frozen in Time~\cite{frozen_in_time} pre-trained on different datasets.}
\label{tab:zero-shot-results-frozen}
\end{table}

\section{Impact of Fusing the Teacher-Student Knowledge}

\Cref{tab:action-recognition-impact,tab:zero-shot-results-impact} present all the metrics for the results on the impact of our method on zero-shot action recognition and zero-shot text-to-video retrieval. Overall, FitCLIP presents the best results. We highlight the importance of fusing the knowledge of the teacher and the student as they individually perform worse than in combination.

\begin{table}
\centering
\subfloat[
\textbf{Moments in Time (MiT)}
\label{tab:mit-results-impact}
]{
\centering
\begin{minipage}{0.9\columnwidth}{\begin{center}
\tablestyle{6pt}{1.0}
\begin{tabular}{crr}
Dataset & Top 1 & Top 5 \\
\shline
Teacher (CLIP) & 19.9 & 40.3 \\
Student & 17.7 & 39.1 \\
\hline
FitCLIP & \textbf{21.8} & \textbf{44.6} \\
$\triangle$ & \textcolor{nice-green}{\(\uparrow\) 1.9} & \textcolor{nice-green}{\(\uparrow\) 4.3} \\
Error rate reduction & \textcolor{nice-green}{\(\uparrow\) 2.4} & \textcolor{nice-green}{\(\uparrow\) 7.2}
\end{tabular}
\end{center}}\end{minipage}
}
\vspace{1em}
\subfloat[
\textbf{UCF101}
\label{tab:ucf101-results-impact}
]{
\centering
\begin{minipage}{0.9\columnwidth}{\begin{center}
\tablestyle{6pt}{1.0}
\begin{tabular}{crr}
Dataset & Top 1 & Top 5 \\
\shline
Teacher (CLIP) & \textbf{74.5} & 94.3 \\
Student & 64.7 & 90.4 \\
\hline
FitCLIP & 73.3 & \textbf{95.3} \\
$\triangle$ & \textcolor{red}{\(\downarrow\) 1.2} & \textcolor{nice-green}{\(\uparrow\) 1.0} \\
Error rate reduction & \textcolor{red}{\(\downarrow\) 4.7} & \textcolor{nice-green}{\(\uparrow\) 17.5}
\end{tabular}
\end{center}}\end{minipage}
}
\caption{\textbf{Impact of fusing teacher-student knowledge on zero-shot action recognition.} $\triangle$ denotes the absolute difference in performance between FitCLIP and the Teacher model.}
\label{tab:action-recognition-impact}
\end{table}

\begin{table}
\centering
\subfloat[
\textbf{MSR-VTT}
\label{tab:msrvtt-results-impact}
]{
\centering
\begin{minipage}{0.9\linewidth}{\begin{center}
\tablestyle{6pt}{1.0}
\begin{tabular}{crrrr}
Dataset & R@1 & R@5 & R@10 & MdR \\
\shline
Teacher (CLIP) & 30.4 & 55.1 & 64.1 & 4 \\
Student & 28.1 & 52.6 & 63.7 & 4 \\
\hline
FitCLIP & \textbf{33.8} & \textbf{59.8} & \textbf{69.4} & \textbf{3} \\
$\triangle$ & \textcolor{nice-green}{\(\uparrow\) 3.4} & \textcolor{nice-green}{\(\uparrow\) 4.7} & \textcolor{nice-green}{\(\uparrow\) 5.3} & \textcolor{nice-green}{\(\uparrow\) 1} \\
Error rate reduction & \textcolor{nice-green}{\(\uparrow\) 4.9} & \textcolor{nice-green}{\(\uparrow\) 10.5} & \textcolor{nice-green}{\(\uparrow\) 14.8} & \textcolor{nice-green}{\(\uparrow\) 25.0\%}
\end{tabular}
\end{center}}\end{minipage}
}
\vspace{1em}
\subfloat[
\textbf{YouCook2}
\label{tab:ycook-results-impact}
]{
\centering
\begin{minipage}{0.9\linewidth}{\begin{center}
\tablestyle{6pt}{1.0}
\begin{tabular}{crrrr}
Dataset & R@1 & R@5 & R@10 & MdR \\
\shline
Teacher (CLIP) & 5.3 & 14.6 & 20.9 & 94 \\
Student & 2.9 & 9.7 & 14.1 & 159 \\
\hline
FitCLIP & \textbf{5.8} & \textbf{15.5} & \textbf{22.1} & \textbf{75} \\
$\triangle$ & \textcolor{nice-green}{\(\uparrow\) 0.5} & \textcolor{nice-green}{\(\uparrow\) 0.9} & \textcolor{nice-green}{\(\uparrow\) 1.2} & \textcolor{nice-green}{\(\uparrow\) 19} \\
Error rate reduction & \textcolor{nice-green}{\(\uparrow\) 0.5} & \textcolor{nice-green}{\(\uparrow\) 1.1} & \textcolor{nice-green}{\(\uparrow\) 1.5} & \textcolor{nice-green}{\(\uparrow\) 20.2\%}
\end{tabular}
\end{center}}\end{minipage}
}
\vspace{1em}
\subfloat[
\textbf{DiDeMo}
\label{tab:didemo-results-impact}
]{
\centering
\begin{minipage}{0.9\linewidth}{\begin{center}
\tablestyle{6pt}{1.0}
\begin{tabular}{crrrr}
Dataset & R@1 & R@5 & R@10 & MdR \\
\shline
Teacher (CLIP) & 26.2 & 49.9 & 60.6 & 5 \\
Student & 20.7 & 42.4 & 54.0 & 8 \\
\hline
FitCLIP & \textbf{28.5} & \textbf{53.7} & \textbf{64.0} & \textbf{4} \\
$\triangle$ & \textcolor{nice-green}{\(\uparrow\) 2.3} & \textcolor{nice-green}{\(\uparrow\) 3.8} & \textcolor{nice-green}{\(\uparrow\) 3.4} & \textcolor{nice-green}{\(\uparrow\) 1} \\
Error rate reduction & \textcolor{nice-green}{\(\uparrow\) 3.1} & \textcolor{nice-green}{\(\uparrow\) 7.6} & \textcolor{nice-green}{\(\uparrow\) 8.6} & \textcolor{nice-green}{\(\uparrow\) 20.0\%}
\end{tabular}
\end{center}}\end{minipage}
}
\caption{\textbf{Impact of fusing teacher-student knowledge on zero-shot text-to-video retrieval.} $\triangle$ denotes the absolute difference in performance between FitCLIP and the Teacher model. To measure the error rate reduction for the median rank, we directly use its reduction rate.}
\label{tab:zero-shot-results-impact}
\end{table}

\section{Alpha Value}

\begin{figure}

\subfloat[
\textbf{Supervised WebVid}
\label{fig:alpha-values-webvid}
]{
\centering
\includegraphics[width=.45\textwidth]{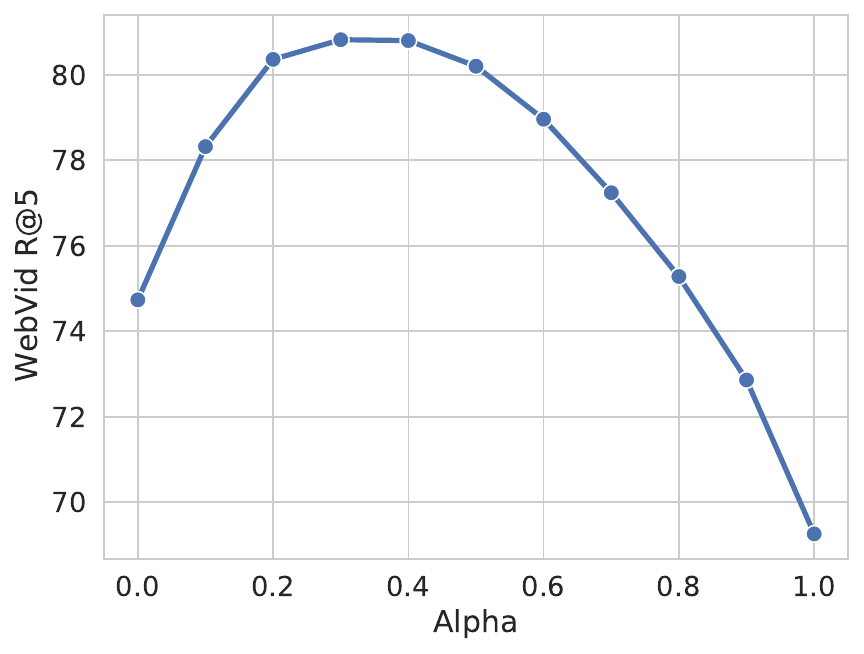}
}
\subfloat[
\textbf{Zero-shot average across 5 datasets}
\label{fig:alpha-values-rest}
]{
\centering
\includegraphics[width=.45\textwidth]{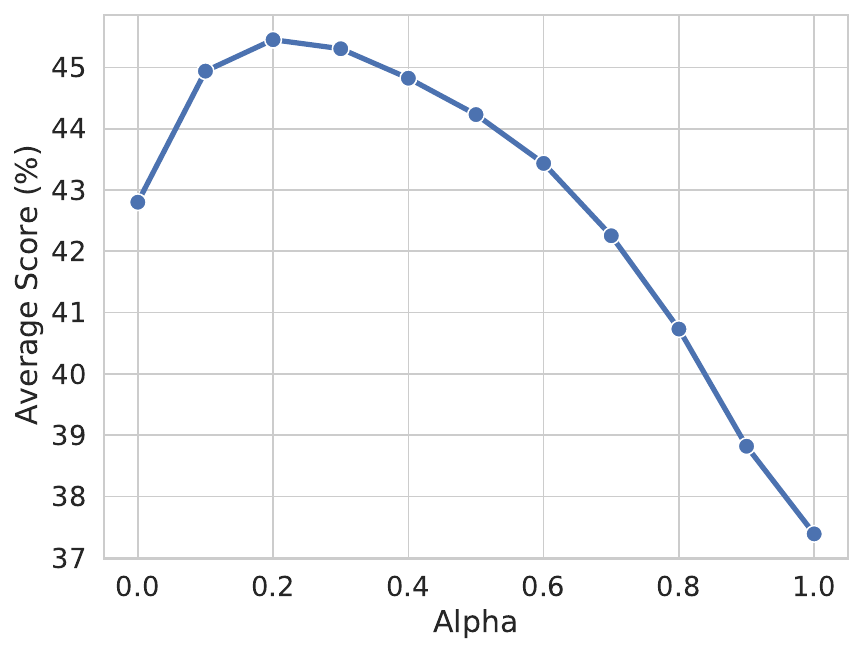}
}

\caption{\textbf{Impact of changing the value of weight-ensembling \(\alpha\) value when fusing the teacher and the student.} We report (a) supervised text-to-video retrieval WebVid R@5 (recall we trained on this domain) and (b) an average across 5 other datasets. The zero-shot text-to-video retrieval datasets used are DiDeMo, MSR-VTT, and YouCook2 (R@5). The zero-shot action recognition datasets used are Moments in Time and UCF-101 (top-1 accuracy). The average value across these datasets is shown.}
\label{fig:alpha-values}
\end{figure}

We analyze the effect of changing the value of \(\alpha\) necessary for the weight-space ensembling step when fusing the teacher and student knowledge in our method. \Cref{fig:alpha-values} shows the effect of this hyperparameter by varying it from 0 to 1, with increments of size 0.1, where 0 is only the teacher and 1 only the student. We show the results on a different split from the training distribution (\cref{fig:alpha-values-webvid}) and on the other datasets we have reported throughout this paper (\cref{fig:alpha-values-rest}).
For WebVid, the best value we obtain is when \(\alpha = 0.3\). Still, we decided to use \(\alpha = 0.4\), which is close enough and the best value obtained by \cite{wortsman2021robust}.
For the other datasets, the best value we obtain is when \(\alpha = 0.2\). For \(\alpha = 0.4\) the score is still high.

\section{Impact of the Labeled Data Size}

\begin{figure}
\centering
\includegraphics{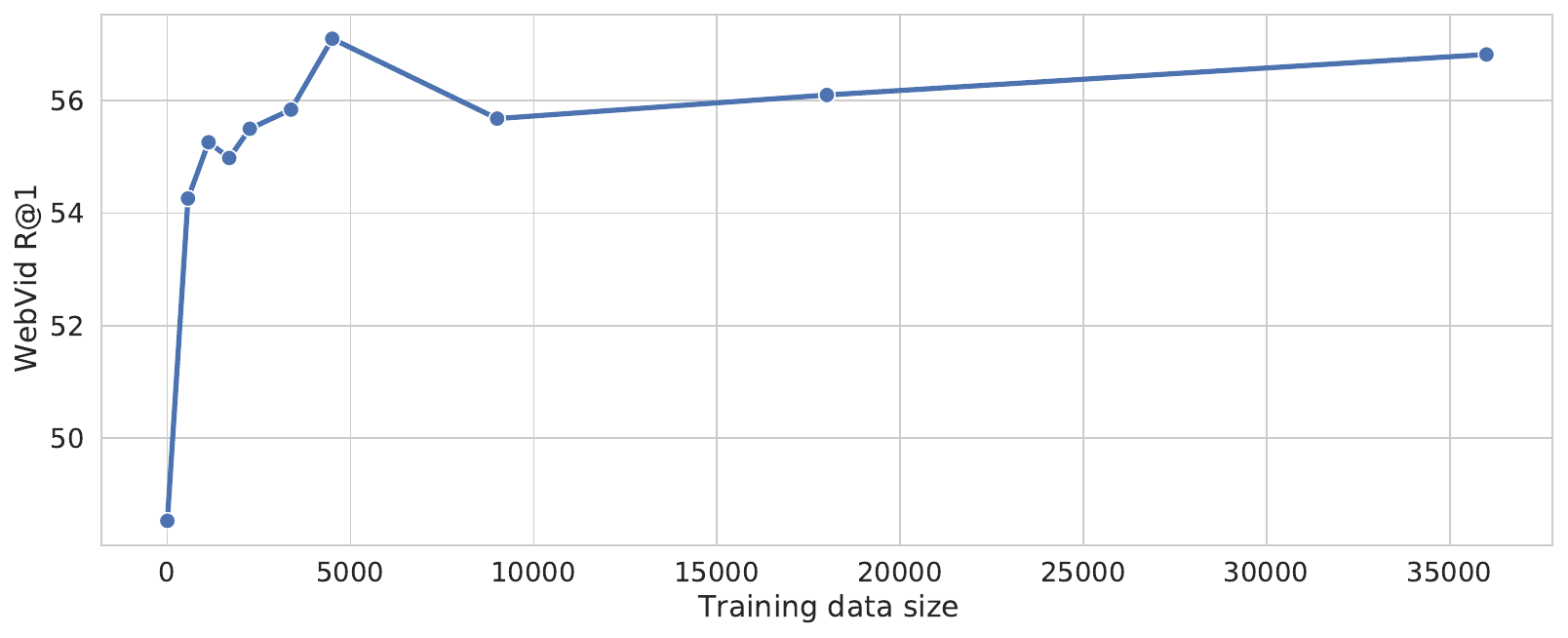}
\caption{\textbf{Text-to-Video top-1 recall on WebVid-2.5M (supervised) of different training subset sizes when fine-tuning CLIP ViT-B/16 and then applying weight-space ensembling.} The evaluated subset sizes are: 0, 563, 1125, 1688, 2250, 3375, 4500, 9000, 18000, and 36000. The subset size 0 represents the evaluation of the pre-trained model without fine-tuning. We exclude large values as we have observed a great drop in performance. Note this experiment doesn't employ distillation.}
\label{fig:data-size}
\end{figure}

The more labeled data for training typically implies the better results. However, more training implies the obtained checkpoint in the weight landscape to be further away from the point of origin and thus harder for weight-ensembling to work well. We study the impact of the labeled data size and try to find a good trade-off point.
\Cref{fig:data-size} show the results of preliminary experiments which are performed by fine-tuning on different subset sizes of the training set from WebVid and applying weight-space ensembling (without distillation).
Each of these subsets where sampled from the whole dataset (they are unlikely subsets of each other).
We find the best value when the WebVid-2.5M training subset size is 4500.
We recognize that we are indirectly using other parts of WebVid, which can boost the in-distribution performance of the selected subset. However, note this doesn't imply better out-of-distribution performance.
We skip showing results for large values as we have observed a great drop in performance. In particular, we obtained results that are considerably worse than the pre-trained model when using the whole training set (2.5M).

\section{Share of Pseudo-Labels/Labels}

\begin{figure}

\subfloat[
\textbf{Moments in Time}
\label{fig:lambda-values-mit}
]{
\centering
\includegraphics[width=.3\textwidth]{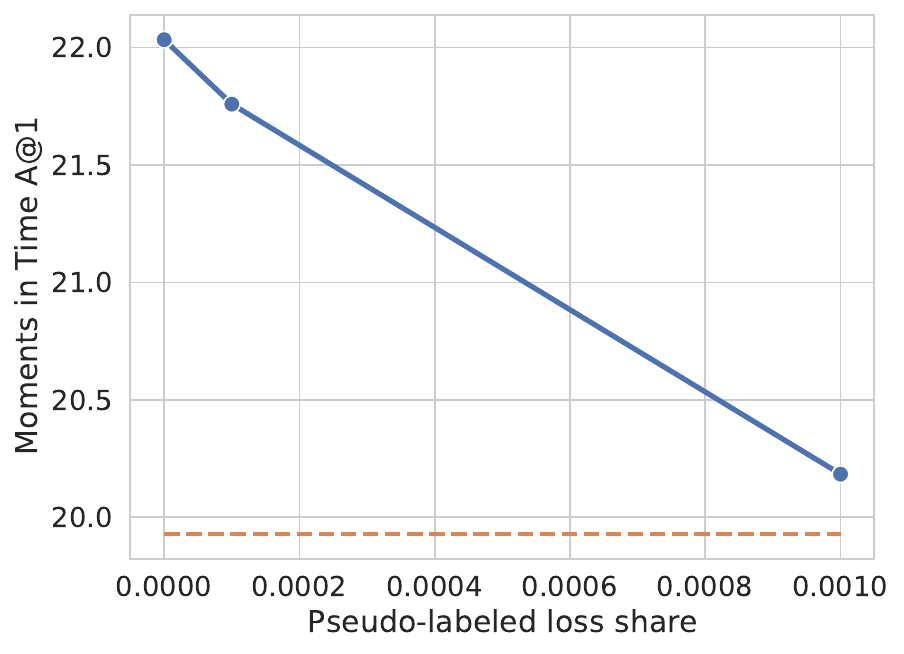}
}
\subfloat[
\textbf{UCF101}
\label{fig:lambda-values-ucf-101}
]{
\centering
\includegraphics[width=.3\textwidth]{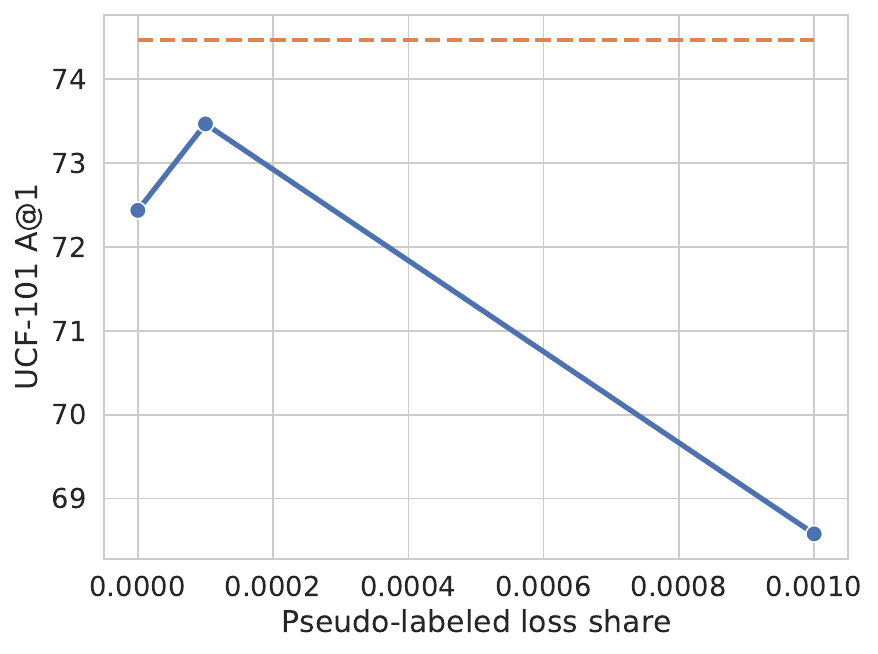}
}
\vspace{1em}
\subfloat[
\textbf{MSR-VTT}
\label{fig:lambda-values-msr-vtt}
]{
\centering
\includegraphics[width=.3\textwidth]{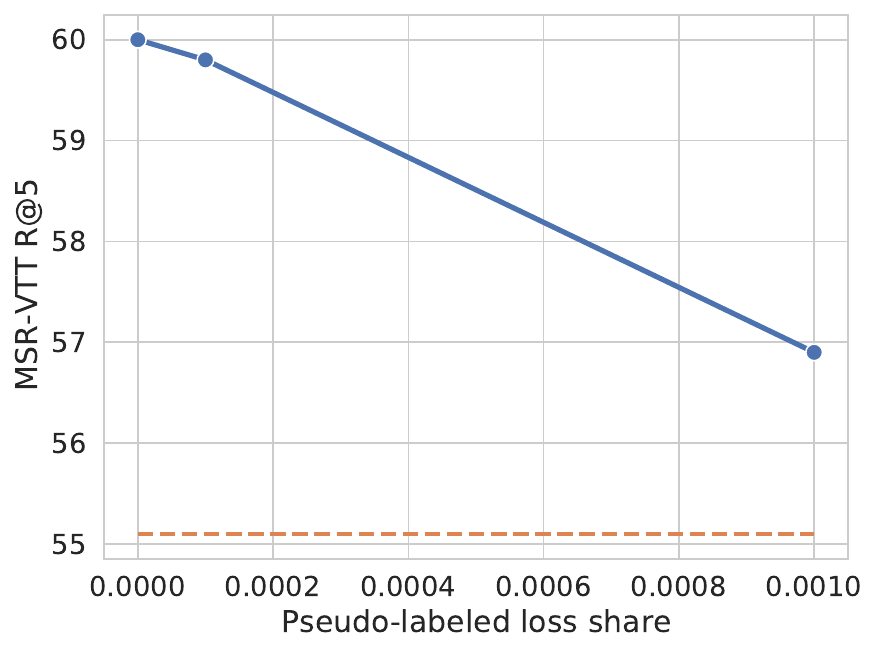}
}
\subfloat[
\textbf{YouCook2}
\label{fig:lambda-values-youcook2}
]{
\centering
\includegraphics[width=.3\textwidth]{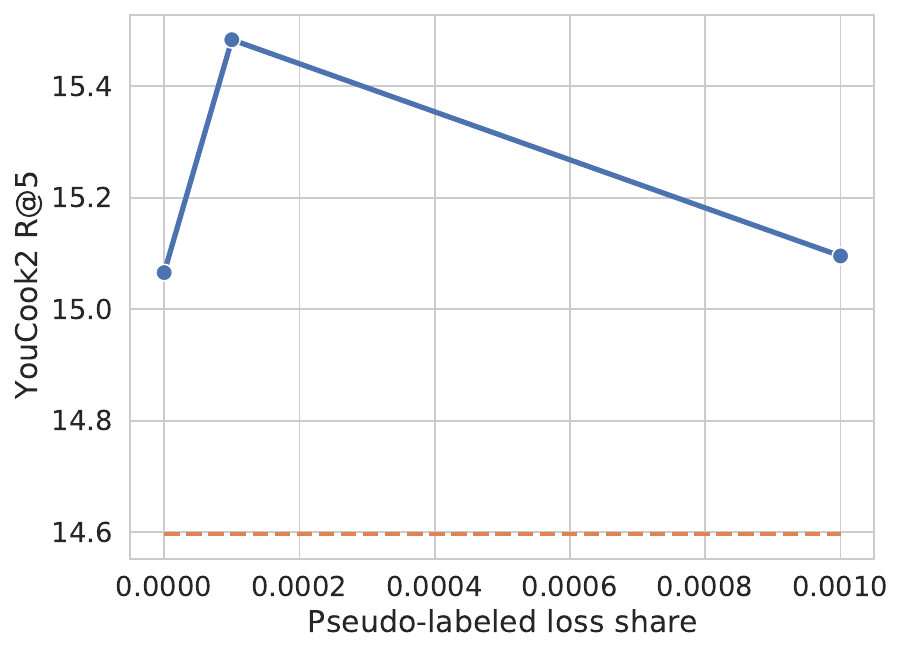}
}
\subfloat[
\textbf{DiDeMo}
\label{fig:lambda-values-didemo}
]{
\centering
\includegraphics[width=.3\textwidth]{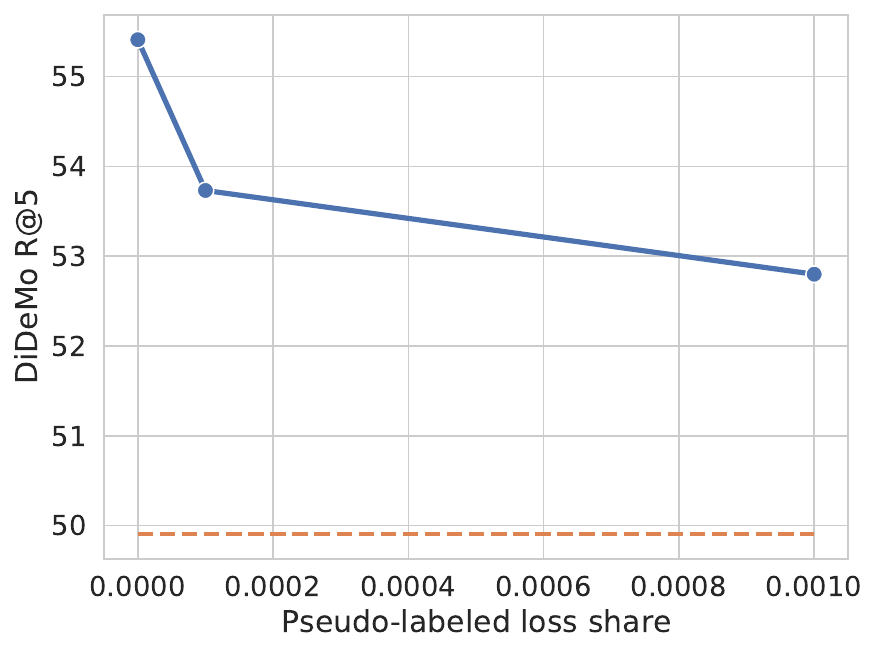}
}

\caption{\textbf{The effect on the zero-shot performance of the share of the pseudo-labeled and labeled losses in FitCLIP.} Each plot shows how the proportion of the pseudo-labeled loss (x-axis) affects the zero-shot performance on a given dataset. The dashed orange line shows the performance of CLIP, as a reference. We skip the sampled values greater than 0.01 to better visualize the plots since they tend to bring a worse performance.
}
\label{fig:lambda-values}
\end{figure}

We are interested in comparing the effect of applying weight-ensembling to a distilled model with applying it to a model that has been trained only on labeled data.
\Cref{fig:lambda-values} shows the effect of varying the proportion of the labeled loss in the final loss in our zero-shot benchmarks.
The use of the distillation loss with \(\lambda={10}^{-4}\) outperforms the usage of only the labeled loss in YouCook2 and UCF101 and shows similar performance on MSR-VTT. In contrast, The performance on DiDeMo and Moments in Time seems to be better with using only the labeled loss.
We hypothesize our method is especially useful on datasets whose distribution (\eg{}, YouCook2) is more distant from the training-time dataset (WebVid-2.5M).

\begin{table}
\centering
\tablestyle{3.5pt}{1.05}
\begin{tabular}{c|rr|rrr}
& \multicolumn{2}{c|}{Action Recognition} & \multicolumn{3}{c}{Text-to-video Retrieval} \\
& UCF101 & MiT & MSR-VTT & YouCook2 & DiDeMo \\
\shline
CLIP    & 74.5          & 19.9          & 55.1          & 14.6          & 49.9 \\
\hline
WiSE-FT & 72.5          & \textbf{22.0} & \textbf{59.9} & 15.1          & \textbf{55.4} \\
FitCLIP & \textbf{73.3} & 21.8          & 59.8          & \textbf{15.5} & 53.7 \\
\end{tabular}
\caption{\textbf{Importance of the Pseudo-Labels.} We report the top-1 accuracy for the zero-shot action recognition datasets, and the top-5 recall for the zero-shot text-to-video retrieval ones. We show in bold the best results between WiSE-FT and FitCLIP for each dataset.}
\label{tab:pseudo-labels}
\end{table}

\bibliography{egbib}